\documentclass[letterpaper]{article} 
\usepackage{aaai24}  
\usepackage{times}  
\usepackage{helvet}  
\usepackage{courier}  
\usepackage[hyphens]{url}  
\usepackage{graphicx} 
\usepackage{natbib}  
\usepackage{caption} 
\usepackage{algorithm}
\usepackage{algorithmic}
\usepackage{multirow}
\usepackage{xspace}
\usepackage{threeparttable}
\usepackage{booktabs}
\usepackage{amssymb}

\makeatletter
\DeclareRobustCommand\onedot{\futurelet\@let@token\@onedot}
\def\@onedot{\ifx\@let@token.\else.\null\fi\xspace}
 
\def\ie{\emph{i.e}\onedot} 
 
\def\etc{\emph{etc}\onedot}

\usepackage{newfloat}
\usepackage{listings}
\DeclareCaptionStyle{ruled}{labelfont=normalfont,labelsep=colon,strut=off} 
\floatstyle{ruled}
\newfloat{listing}{tb}{lst}{}
\floatname{listing}{Listing}
\title{Relevant Intrinsic Feature Enhancement Network\\
for Few-Shot Semantic Segmentation}
\author{
    Xiaoyi Bao,\equalcontrib\textsuperscript{\rm 1,\rm 2,\rm 3}
    Jie Qin,\equalcontrib\textsuperscript{\rm 1,\rm 2}
    Siyang Sun,\textsuperscript{\rm 3}
    Yun Zheng,\textsuperscript{\rm 3}
    Xingang Wang\textsuperscript{\rm 2}\thanks{Corresponding author.}
}
\affiliations{
    \textsuperscript{\rm 1}School of Artificial Intelligence, University of Chinese Academy of Sciences\\
    \textsuperscript{\rm 2}Institute of Automation, Chinese Academy of Sciences\
    \textsuperscript{\rm 3}Alibaba group\\
    \{baoxiaoyi2021, qinjie2019\}@ia.ac.cn,
    \{siyang.ssy,zhengyun.zy\}@alibaba-inc.com,
    xingang.wang@ia.ac.cn
}

\begin{document}

\maketitle

\begin{abstract}
For few-shot semantic segmentation, the primary task is to extract class-specific intrinsic information from limited labeled data. However, the semantic ambiguity and inter-class similarity of previous methods limit the accuracy of pixel-level foreground-background classification. To alleviate these issues, we propose the Relevant Intrinsic Feature Enhancement Network (RiFeNet). To improve the semantic consistency of foreground instances, we propose an unlabeled branch as an efficient data utilization method, which teaches the model how to extract intrinsic features robust to intra-class differences. Notably, during testing, the proposed unlabeled branch is excluded without extra unlabeled data and computation. Furthermore, we extend the inter-class variability between foreground and background by proposing a novel multi-level prototype generation and interaction module. The different-grained complementarity between global and local prototypes allows for better distinction between similar categories. The qualitative and quantitative performance of RiFeNet surpasses the state-of-the-art methods on $PASCAL-5^i$ and $COCO$ benchmarks.

\end{abstract}

\begin{figure}[!t]
\centering
   \includegraphics[width=\linewidth]{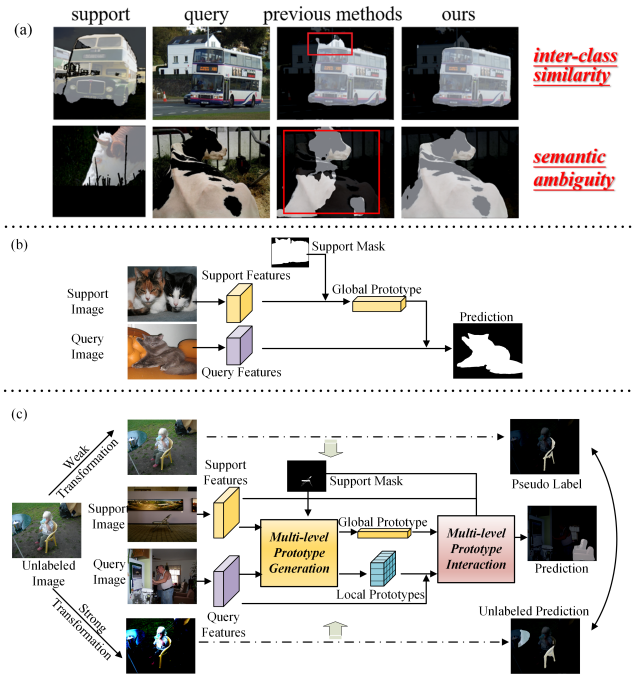}

   \caption{Comparison of RiFeNet and other works. (a) Two foreground-related issues that limit the effectiveness of previous research. (b) Schedule of previous prototype-based methods. The two-branch structure uses a single global prototype as the only medium for information interaction. (c) Framework of RiFeNet. An additional unlabeled branch is used for feature mining during training, with multi-granularity prototypes extracted from different branches. }
\label{introduction}
\end{figure}

\section{Introduction}
As a fundamental and crucial task in the field of computer vision, semantic segmentation is widely applied in visual tasks such as medical image understanding, industrial defect inspection, virtual reality games, autonomous driving, \etc. With the development of convolutional neural networks, fully-supervised semantic segmentation has achieved remarkable success \cite{long2015fully}. Subsequent transformer-based methods also greatly improve the segmentation performance \cite{zhu2020deformable,xie2021segformer,yuan2021hrformer}. However, it is still arduous to acquire pixel-level annotations that require a huge amount of manual effort and costs. To alleviate such expensive and unwieldy annotations, many works tend to resort to a few-shot semantic segmentation paradigm. In this setting, the model learns to segment with a few limited labeled data as the support set and then transferred to test the query inputs. 

Previous approaches \cite{shaban2017one,liu2020part} focus on making full use of limited support data. The training data is divided into a support set and a query set, which are processed using two different branches. Relationship prompt and information interaction between the two branches are achieved by proposing superior matching strategies \cite{kang2022integrative} or generating representative prototypes \cite{tian2020prior,wang2019panet,zhang2019canet,zhang2020sg}. The former class of methods is represented by HSNet \cite{min2021hypercorrelation}, which usually uses a cost volume method to aggregate dense correlation scores in high-dimensional space by 4D convolution. Subsequent researches improve the network using a swin transformer \cite{hong2022cost} or a doubly deformable transformer \cite{xiong2022doubly} in 4-dimension space. With much lower computational costs, prototype-based methods also achieve good segmentation results. As illustrated in Fig.~\ref{introduction} (a), the single or multiple prototypes integrating semantic class-specific features are extracted from the support branch and used to activate query features. Researchers have proposed different activation methods in PPNet \cite{liu2020part}, PFENet \cite{tian2020prior}, CyCTR \cite{zhang2021few}, \etc, to fully exploit and utilize the limited number of category semantic features.

The object to be segmented is called the foreground while the background refers to the rest of the stuff and things. Despite the prevailing success of these methods, there are still two main issues affecting the segmentation effect of previous methods, \ie \textbf{semantic ambiguity} and \textbf{inter-class similarity}, as shown in Fig.\ref{introduction} (a). For the foreground itself, semantic ambiguity appears for different instances of the same class. The intra-class differences between the two branches lead to semantic mistakes on query images. Further to the problem of distinguishing between foreground and background, inter-class similarity is specifically the difficulty in pixel-level binary classification. When objects of different classes with similar textures appear simultaneously, the local features of foreground and background become confusing.

The semantic ambiguity of the foreground is caused by the poor intra-class generalization of the model. Previous models mainly explore how to match the support set and the query set. With little labeled data in few-shot tasks, their models are prone to extract semantically ambiguous shallow features such as shape and pose, and thus unable to learn robust class representations for highly diverse query appearances. This distribution discrepancy makes the model identify the semantic parts of the foreground objects incompletely and segment them inappropriately. As for the inter-class similarity between foreground and background, the lack of discriminative features for query data is to blame for this problem. Specifically, previous methods extract global prototypes or prototypes with a single level of granularity. The information they carry is inadequate and monolithic, thus leading to pixel misidentify.

\begin{figure*}[t]
\centering
\includegraphics[width=0.9\linewidth]{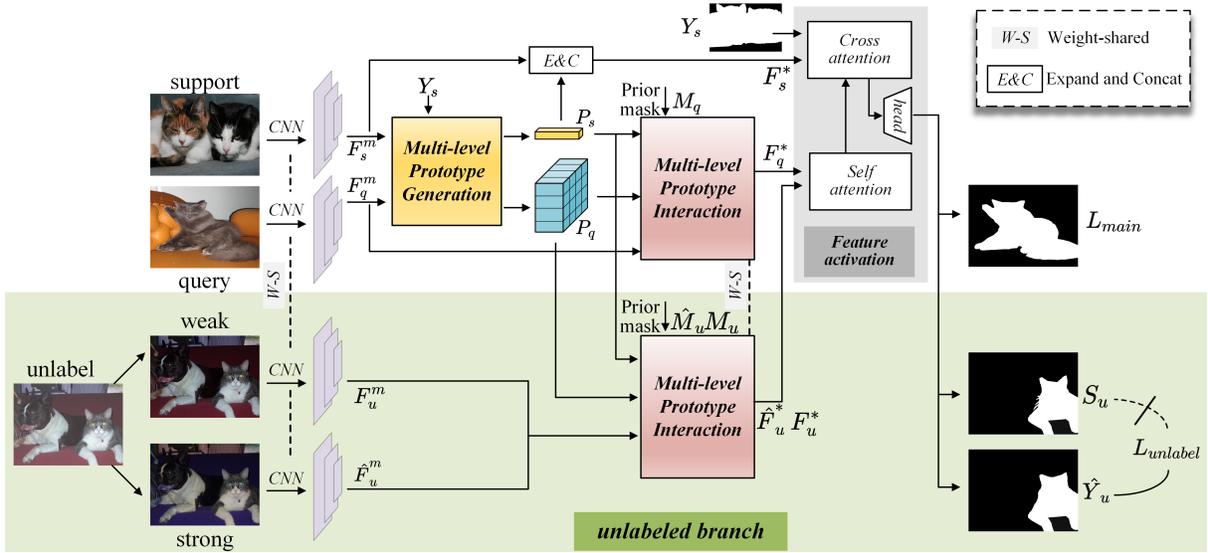}

   \caption{Overall architecture of RiFeNet. An additional branch using unlabeled inputs is attached to the traditional two-branch structure, as shown in green. The forward process of RiFeNet consists of three main blocks. The multi-level prototype generation block generates global and local prototypes. The multi-level prototype interaction module allows for interaction and integration between prototypes. A feature activation block is used consequently for obtaining final segmentation results.}
\label{overall}
\end{figure*}

To address the above problems, we propose a novel relevant intrinsic feature enhancement network, as shown in Fig.~\ref{introduction} (c). For foreground objects, we enhance the semantic consistency of the same class, thus improving the intra-class generalization of features. We incorporate a novel unlabeled branch into the support-query pair branches to teach the model how to mine intrinsic robustness behind appearance variability. The additional unlabeled data branch serves as a regularization term during training that enhances the semantic consistency of few-shot segmentation models. 

We go one step further in achieving an overall enhancement of the distinction between foreground and background. A novel multi-level prototype generation and interaction module is proposed to ensure inter-class variability. The multi-level prototype features contain a global prototype from the support data and a local prototype from the query data. The former represents the high-level semantic abstraction about the entire structure of the target category, while the latter captures the fine-grained appearance concepts providing details for category discrimination. The global and local prototype interaction can complement each other to ensure that the network extracts the corresponding category features. This multi-level prototype feature interaction module can greatly widen the inter-class differences, benefiting the identification between foreground and background. 

Our contributions can be summarized as follows:
\begin{itemize}
   \item We propose a relevant intrinsic feature enhancement network. By alleviating semantic ambiguity and inter-class similarity, our model improves the forehead segmentation performance in this few-shot task.
   \item To maintain the foreground semantic consistency, we propose an unlabeled branch as an efficient data utilization method, which teaches the model how to extract intrinsic features robust to intra-class differences. 
   \item To further achieve an overall effective distinction between foreground and background objects, we propose a novel multi-level prototype generation and interaction module to extend the inter-class variability. With different-grained semantic information from different sources, the mutual complementarity of information is facilitated and the discriminative representation ability of confusing foreground and background objects is boosted.
   \item Extensive experiments demonstrate the effectiveness of our proposed method, which achieves state-of-the-art accuracy on both PASCAL-$5^i$ and COCO benchmarks.
\end{itemize}


\section{Related Work}
\subsection{Semantic Segmentation}
Rapid progress has been made on semantic segmentation tasks since fully convolutional networks (FCN) transformed them into pixel-level classification \cite{long2015fully}. After that, the corresponding encoder-decoder architecture has been widely used \cite{qin2022activation,qin2022multi,qin2023freeseg}. Recent research mainly focuses on multi-scale feature fusion \cite{zhao2017pyramid,chen2018encoder,yang2018denseaspp,he2019adaptive}, insertion of attention modules \cite{fu2019dual,yuan2018ocnet,li2019expectation,tao2020hierarchical,zhu2019asymmetric,zhang2019acfnet}, and context prior \cite{lin2017refinenet,zhang2018context,yu2020context,jin2021mining}. 

Inspired by the success of the vision transformer \cite{dosovitskiy2020image}, researchers have carried out attempts to apply the transformer structure to segmentation \cite{yuan2021hrformer,wang2021pyramid,lee2022mpvit,xie2021segformer,strudel2021segmenter}. These works perform well under the task setting of semantic segmentation. However, in practical application scenarios, they are unable to cope with sparse training data and thus often fail on categories that they have not seen during the training process.

\subsection{Few-Shot Segmentation}
A two-branch structure is widely used in few-shot segmentation, \ie, a support branch and a query branch. Mainstream few-shot methods are divided into two categories: prototype extraction and spatial correlation. With the development of image-matching, spatial correlation-based methods have regained attention in few-shot segmentation. HSNet \cite{min2021hypercorrelation} applies 4D convolution to 4D correlation tensors in a pyramid structure, and the following ASNet \cite{kang2022integrative} squeezes semantic correlation into the foreground map. VAT \cite{hong2022cost} and DACM \cite{xiong2022doubly} use cost volume aggregation-based models with 4D transformers.  

The above methods benefit from the retained spatial structure. However, their high computational complexity and the excessive number of parameters are the drawbacks, making training slow and failing to generalize. Since PL \cite{dong2018few} introduced prototype learning to few-shot segmentation, most of the following research focus on prototype-based methods \cite{zhang2019canet,wang2019panet}. SG-ONE \cite{zhang2020sg} proposes to use masked average pooling to extract a prototype from support features carrying the category information. Dense matching computation is performed between query features and this prototype. Tian proposes the PFENet \cite{tian2020prior}, adding a train-free prior mask to the feature enrichment process. Multi-prototypes are generated in PPNet \cite{liu2020part}, ASGNet \cite{li2021adaptive}
, and RPMMs \cite{yang2020prototype} to include more local information with different generating methods, while the use of multiple prototypes in these networks does not bring them more competitive results. IPMT \cite{liu2022intermediate} proposes an intermediate prototype and uses the transformer to update the prototype iteratively. 

SSP \cite{fan2022self} also proposes the idea of the self-support prototype, but their single spacial-agnostic prototype is generated for self-matching. Although also generated from query branches, our multiple local prototypes carry discriminative and comprehensive spatial information that helps to provide details for intra-class discrimination.

\subsection{Unlabeled Usage in Few-Shot Segmentation}
Very few researchers have explored the leverage of unlabeled data in few-shot tasks. PPNet \cite{liu2020part} is one of the cases, which incorporates 100 unlabeled data into every support input, with the use of a graph neural network (GNN). Based on it, Soopil proposes an uncertainty-aware model to make more adequate use of unlabeled data without using graph neural networks \cite{kim2023uncertainty}. However, the above two methods also require additional unlabeled data of the current novel class, which is inconsistent with the original few-shot task setting but relevant to the semi-supervised paradigm. And the number of unlabeled images used in every meta-training process far exceeds that of support and query images. In contrast, our work removes the dependence of the testing process on unlabeled data. During training, a tiny amount of unlabeled data is used to constrain the foreground semantic similarity.

\section{Method}
\subsection{Problem Definition}
The few-shot semantic segmentation task is defined as the segmentation of novel-class objects, based on a very small number of images with pixel-level annotations. According to this definition, the dataset for training $D_{train}$ and the one for testing $D_{test}$ have no overlap in their categories. 

In each meta-training process of $K$-shot setting, conventional methods randomly sample $K+1$ image pairs of the same class $j$: $\{(I^i,Y^i),i\in\{1,2,...K+1\} |(I^i,Y^i)\in D_{train},\;C^i=j\}$. $ I^i $, $Y^i$, and $C^i$ refer to the RGB image, its segmentation mask, and the class of the segmented object. Among them, $K$ pairs are sent to the support branch, and the last is used as the query input and ground truth. The difference in our network is that we take $K+1+M$ pairs of images from $D_{train}$ at a time, and $I^i$ of the extra $M$ pairs are used as input to the unlabeled branch. 

In each testing process, the $K+1$ pairs from the $j$-th novel classes$\{(I^i,Y^i),i\in\{1,2,...K+1\} |(I^i,Y^i)\in D_{test},\;C^i=j\}$ are sampled and assigned to support and query branch as before.

\subsection{Relevant Intrinsic Feature Enhancement}
As a pixel-level binary classification task, semantic segmentation needs to identify the foreground, \ie{object to be segmented}, and the background, \ie{the rest of the stuff and things}. To alleviate the semantic ambiguity and inter-class similarity in Fig.\ref{introduction}, we propose this relevant intrinsic feature enhancement network. We present the complete framework under the 1-shot setting in Fig.~\ref{overall}. The proposed RiFeNet consists of three branches with a shared backbone. The extra unlabeled branch helps the traditional support-query framework learn how to assure semantic consistency. 

The forward process consists of three main modules: the multi-level prototype generation module, the multi-level prototype interaction module, and the feature activation module. The first two modules are used to provide multi-grained evidence for better inter-class discrimination. The feature activation module activates pixels containing objects of the target class and deactivates the others, providing the final segmentation result.

\subsection{Feature Enhancement with Unlabeled Data}
There are large intra-class distributional differences in the foreground objects that need to be segmented. Therefore, the training purpose is to ensure
the semantic consistency of features extracted from different instances of the foreground category, rather than focusing only on semantically ambiguous appearance features.

Therefore, we introduce an auxiliary unlabeled branch as an effective data utilization method to aid model learning. Without adding training data, we re-sample a subset of training samples as unlabeled data, with the same segmentation loss applied. By augmenting the sample diversity, it teaches the model to avoid learning sample-specific biases of labeled inputs, even in the absence of unlabeled data during testing.

Our unlabeled branch shares parameters with the query branch to align features and enhance relevance. It is trained with pseudo-labels generated from each other. Specifically, the initial unlabeled input goes through two versions of data augmentation, different in their kinds and intensities. We use $I_u$ and $\hat{I}_u$ to refer to the weakly and strongly transformed unlabeled input. Both are sent to the backbone simultaneously with an identical forward process, thus we use symbols without a hat in all subsequent equations for simplicity.
\begin{equation}
\label{pre-process}
	F^{m}_u = ReLU(Conv(CAT(F^{1}(I_u),F^{2}(I_u)))). 
\end{equation}%
$F^{m}_u$ refers to merged unlabeled feature maps. For the sake of simplicity, we omit the above processing in Fig.\ref{overall}. The support and query branches are handled in the same way, and all parameters in Eq.\ref{pre-process} are shared with query branches. $F^j(I_u)$ corresponds to the feature map from the $j$-th backbone layer.
\begin{figure}
\centering
   \includegraphics[width=\linewidth]{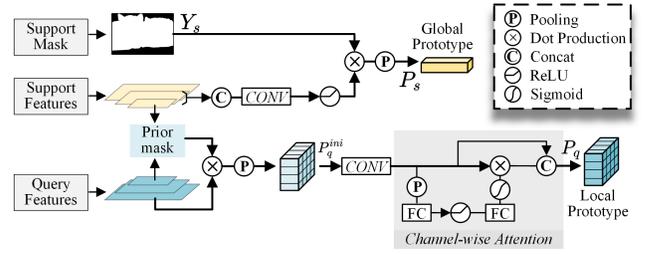}

   \caption{Visual illustration of the multi-level prototype generation block that extracts global and local prototypes. }
\label{generation}
\end{figure}

As for the prototype generation block, if we follow the query paradigm to generate local prototypes from the unlabeled branch, the self-supervised training may be off-target in a few extreme cases. In complex environments with multiple objects, the segmented class in the generated pseudo mask may not be inconsistent. This off-track pseudo-truth may worsen the expression capability of the model. 

We introduce query prototypes here to provide class-specific prior. Only intrinsic features with intra-class relevance are extracted when processing unlabeled data. 

The subsequent prototype interaction is the same as the query one. Augmented unlabeled features are obtained and sent to the activation block for prediction. $S^u$ and $\hat{Y}^u$ represent the predicted results of weakly and strongly transformed inputs, respectively. The former output $S^u$ serves as a pseudo label for the latter one. The loss of this self-supervised process is calculated as follows, with dice loss (\citeauthor{milletari2016v}) as its training objective.
\begin{equation}
            \mathcal{L}_{unlabel} = DICE(S^u,\hat{Y}^u).
\end{equation}%

\begin{figure}
\centering
   \includegraphics[width=\linewidth]{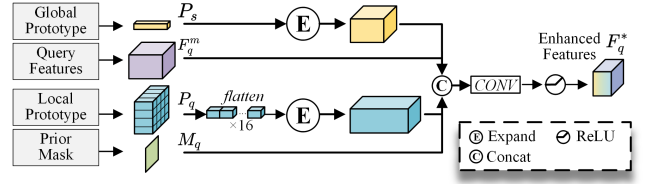}

   \caption{Schedule of the multi-level prototype interaction merging relevant intrinsic information for enhancement. }
\label{interaction}
\end{figure}

\subsection{Multi-level Prototype Processing} 
While the unlabeled branch ensures intra-class semantic consistency of the foreground, the background noise it introduces may worsen existing inter-class similarity between foreground and background. Therefore, we propose the multi-level prototype generation and interaction blocks aim to capture representative information in different granularity and augment the discrimination of foreground and background in confusing areas.

\textbf{Global support prototype generation.} 
To capture category information at the high-dimensional category level, we extract global prototypes from support features.

As shown in Fig.\ref{generation}, $P_{s} \in \mathbb{R}^{1\times 1\times C}$ is obtained through masked average pooling on the global feature. $w$ and $h$ are the width and height of $F^{m}_s$. $\odot$ means dot production. 
\begin{equation}
            P_{s} = \frac{\sum^{w,h}_{x=1,y=1}(F^{m}_s \odot Y_s)^{x,y}}{\sum^{w,h}_{x=1,y=1}Y_{s}^{x,y}}.
\end{equation}%

\textbf{Local query prototype generation.} 
In some cases, the similarities between different classes may be more significant than the differences within classes \cite{fan2022self}. Meanwhile, the prototypes obtained by global average pooling completely discard spatial structure information, making the decoupling of local components unrealistic.

To provide discriminative details, we additionally extract local prototypes from the query branch. These prototypes facilitate binary classification by prompting fine-grained information. Fig.\ref{generation} illustrates their generation process. 
\begin{equation}
            P^*_{q} = Pooling_{avg}(F^{m}_q \odot M(F^3(I_q))).
\end{equation}%
$M(F^3(I_q))$ is the prior mask proposed in PFENet \cite{tian2020prior}, as a form of similarity measure between high-level feature maps of support and query branches. Local average pooling is done on the multiplication of prior masks and query feature maps. 

To squeeze the channel and enhance the representation of local prototypes, $1\times1$ convolution and channel-wise attention are used. The refined local prototypes $P_{q} \in \mathbb{R}^{m\times m\times C'}$ from the query branch itself are obtained as follows. The symbol $Att_{chn}$ refers to channel-wise attention.
\begin{equation}
            P_{q} = Att_{chn}(Conv^{1\times 1}(P^*_{q})).
\end{equation}%

For the generation of prior masks, fixed high-level feature maps $F^{4}_{s}$ and $F^{4}_{q}$ are more suitable. But when generating local query prototypes, superficial partly appearance information counts a lot. With the need for pixel-level correspondence, we use feature map $F^{1}_{q}$ from the first layer to extract locally consistent appearance information in the local area, as shown in Fig.\ref{generation}.

\textbf{Multi-level prototype interaction.} 
To emphasize the intrinsic class information, we build interactions between different-grained prototypes, thus enhancing the feature-mining ability for identification. 

Fig.\ref{interaction} shows detailed steps of our interaction block. The generated global and local prototypes are expanded to the size of feature maps and then concatenated with query features and the prior mask. After a $1\times1$ convolution with activation, augmented query features $F^{*}_q$ are obtained.
\begin{equation}
	 F^{*}_q = ReLU(Conv(CAT(P_s,P_q,F^{m}_q,M(F^3(I_q))))). 
\end{equation}%

\subsection{Feature activation}
For deeper information interaction between the query and the support branch, we use an n-layer transformer encoder for feature activation. In this process, information with relevance is activated, enhancing the total feature map.

Augmented query features $F^{*}_q$ are processed by self-attention and then cross-attention with support features $F^{*}_s$. Each block is made up of a multi-head attention and a feed-forward network. The setting of multi-head attention for cross-attention follows the one in CyCTR \cite{zhang2021few}, with the segmentation mask of the support image $Y_s$ used to maintain the cycle consistency.
\begin{equation}
	F^{final}_q = Att_{cross}(Att_{self}(F^{*}_q),F^{*}_s,Y_s). 
\end{equation}%
$Att_{cross}$ and $Att_{self}$ represent the cross-attention and self-attention block, respectively. The output of the transformer $F^{final}_q$ is resized and passed to the classification head to obtain the final pixel-by-pixel segmentation result, as shown in the following equation. 
\begin{equation}
	Y_q = CLS(F^{final}_q+ReLU(Conv(CAT(F^{final}_q))). 
\end{equation}%
$Y^q$ represents the predicted result of the query input and $CLS$ refers to the segmentation head.

The loss of RiFeNet is calculated between predictions of query inputs and their corresponding ground truth. We choose dice loss as the loss function, thus the loss for each meta-learning task can be represented as:
\begin{equation}
            \mathcal{L}_{main} = DICE (\hat{Y}^q,Y^q) + \mathcal{L}_{aux}.
\end{equation}%
where $\hat{Y}^q$ and ${Y}^q$ are the prediction and ground truth of query images. $\mathcal{L}_{aux}$ refers to an auxiliary loss attached for feature alignment, following the common setting \cite{milletari2016v}.
\begin{equation}
            \mathcal{L}_{final} = \mathcal{L}_{main} + \beta \mathcal{L}_{unlabel}.
\end{equation}%

As the above equation shows, the final loss is a weighted sum of the main loss and the self-supervised loss $\mathcal{L}_{unlabel}$. The weight $\beta$ is set to $0.5$ empirically.


\begin{table*}[!t]
\centering
\begin{threeparttable}
\resizebox{0.9\linewidth}{!}{
\begin{tabular}{l|c|ccccc|ccccc|c}
\toprule[1pt]
\multirow{2}{*}{Method}&\multirow{2}{*}{Backbone} & \multicolumn{5}{c|}{1-shot}& \multicolumn{5}{c|}{5-shot} &\#learnable\\
& &split0 & split1 & split2 & split3 & mean 	& split0 & split1 & split2 & split3 & mean & params\\
\midrule
PPNet~\citeyear{liu2020part} 		& \multirow{9}{*}{Res-50} 	&52.7&62.8&57.4&47.7&55.2	&60.3&70.0&69.4&60.7&65.1	&31.5M\\
PMM~\citeyear{yang2020prototype} &&52.0&67.5&51.5&49.8&55.2&55.0&68.2&52.9&51.1&56.8&-\\
PFENet~\citeyear{tian2020prior}   	& 						 	&61.7&69.5&55.4&56.3&60.8	&63.1&70.7&55.8&57.9&61.9	&10.8M\\
CyCTR~\citeyear{zhang2021few}   &							&65.7&71.0&59.5&\underline{59.7}&64.0	&69.3&73.5&63.8&63.5&67.5	&7.4M\\
HSNet~\citeyear{tao2020hierarchical}   	& 							&64.3&70.7&60.3&\textbf{60.5}&64.0	&\underline{70.3}&73.2&67.4&\textbf{67.1}&\underline{69.5}	&2.6M\\
ASGNet~\citeyear{li2021adaptive}		&							&58.8&67.9&56.8&53.7&59.3	&63.7&70.6&64.1&57.4&63.9	&10.4M\\
SSP~\citeyear{fan2022self}		&							&61.4&67.2&\underline{65.4}&49.7&60.9	&68.0&72.0&\textbf{74.8}&60.2&68.8	&-\\
DCAMA~\citeyear{shi2022dense}		&							&\underline{67.5}&\underline{72.3}&59.6&59.0&\underline{64.6}	&\textbf{70.5}&\underline{73.9}&63.7&\underline{65.8}&68.5	&-\\ 
\textbf{RiFeNet (Ours)}&							&\textbf{68.4}&\textbf{73.5}&\textbf{67.1}&59.4&\textbf{67.1}	&70.0&\textbf{74.7}&\underline{69.4}&64.2&\textbf{69.6}	&7.7M\\
\midrule
DAN~\citeyear{wang2020few}       	&\multirow{9}{*}{Res-101}	&54.7&68.6&57.8&51.6&58.2	&57.9&69.0&60.1&54.9&60.5	&-\\
PMM~\citeyear{yang2020prototype} &&54.7&68.6&57.8&51.6&58.2 &57.9&69.0&60.1&54.9&60.5&-\\
PFENet~\citeyear{tian2020prior}   	& 						 	&60.5&69.4&54.4&55.9&60.1	&62.8&70.4&54.9&57.6&61.4	&10.8M\\
CyCTR~\citeyear{zhang2021few}  	&							&67.2&71.1&57.6&59.0&63.7 &\underline{71.0}&\underline{75.0}&58.5&65.0&67.4	&7.4M\\
HSNet~\citeyear{tao2020hierarchical}    	& 							&\underline{67.3}&\underline{72.3}&62.0&\textbf{63.1}&\underline{66.2}	&\textbf{71.8}&74.4&67.0&\textbf{68.3}&\underline{70.4}	&2.6M\\
ASGNet~\citeyear{li2021adaptive}		&							&59.8&67.4&55.6&54.4&59.3	&64.6&71.3&64.2&57.3&64.4&10.4M\\
SSP~\citeyear{fan2022self}		&							&63.7&70.1&\textbf{66.7}&55.4&64.0  &70.3&\textbf{76.3}&\textbf{77.8}&\underline{65.5}&\textbf{72.5}&-	\\
DCAMA~\citeyear{shi2022dense}		&							&65.4&71.4&63.2&58.3&64.6	&70.7&73.7&66.8&61.9&68.3&-	\\ 
\textbf{RiFeNet (Ours)}&							&\textbf{68.9}&\textbf{73.8}&\underline{66.2}&\underline{60.3}&\textbf{67.3}	&70.4&74.5&\underline{68.3}&63.4&69.2&7.7M\\
\bottomrule[1pt]     
\end{tabular}}
\end{threeparttable}
\caption{Performance comparison on PASCAL-$5^i$ in terms of mIoU ($\%$).}
\label{sotapascal}
\end{table*}

\begin{table*}[t]
\centering
\begin{threeparttable}
\resizebox{0.9\linewidth}{!}{
\begin{tabular}{l|c|ccccc|ccccr}
\toprule[1pt]
\multirow{2}{*}{Method}&\multirow{2}{*}{Backbone} & \multicolumn{5}{c|}{1-shot}& \multicolumn{5}{c}{5-shot} \\
& &split0 & split1 & split2 & split3 & mean 	& split0 & split1 & split2 & split3 & mean \\
\midrule
PPNet~\citeyear{liu2020part} 		& \multirow{7}{*}{Res-50} 	&28.1&30.8&29.5&27.7&29.0	&39.0&40.8&37.1&37.3&38.5	\\
PMM~\citeyear{yang2020prototype} &&29.3&34.8&27.1&27.3&29.6 &33.0&40.6&30.3&33.3&34.3\\
RPMMs~\citeyear{yang2020prototype}    &                       &29.5&36.8&28.9&27.0&30.6&33.8&42.0&33.0&33.3&35.5\\
CyCTR~\citeyear{zhang2021few}   & 							&38.9&43.0&39.6&39.8&40.3	&41.1&48.9&45.2&\underline{47.0}&45.6	\\
HSNet~\citeyear{tao2020hierarchical}   	& 							&36.3&43.1&38.7&38.7&39.2	&43.3&\underline{51.3}&48.2&45.0&46.9	\\
SSP~\citeyear{fan2022self}		&			&\textbf{46.4}&35.2&27.3&25.4&33.6&\textbf{53.8}&41.5&36.0&33.7&41.3\\
DCAMA~\citeyear{shi2022dense}		&			&\underline{41.9}&\underline{45.1}&\underline{44.4}&\underline{41.7}&\underline{43.3}&45.9&50.5&\textbf{50.7}&46.0&\underline{48.3}\\
\textbf{RiFeNet (Ours)}&							&39.1&\textbf{47.2}&\textbf{44.6}&\textbf{45.4}&\textbf{44.1} 	&\underline{44.3}&\textbf{52.4}&\underline{49.3}&\textbf{48.4}&\textbf{48.6}	\\
\bottomrule[1pt]     
\end{tabular}}
\end{threeparttable}
\caption{Performance comparison on COCO in terms of mIoU ($\%$).}
\label{sotacoco}
\end{table*}

\section{Experiments}
We conduct experiments on the $PASCAL-5^i$ and $COCO$ datasets. The primary evaluation metric is mean intersection-over-union (mIoU). The foreground-background IoU (FB-IoU) is used as an auxiliary indicator.

\subsection{Process of training and testing}

The number of support: query: unlabeled images is $1:1:2$ or $5:1:2$ for each meta-training task in the one-shot and five-shot settings. The unlabeled input branch comes from the same training dataset. We resample $M$ extra images of the same class without mask truth, rather than adding another dataset. For a fair comparison, we exclude the unlabeled images during testing.

Pre-trained ResNet50 and ResNet101 \cite{he2016deep} are used as the backbone of the model, frozen without parameter update. The baseline method refers to the CyCTR.
Local query prototypes are generated by the multi-level prototype generation. Concatenated with one global support prototype, $m^2$ local query prototype, and the prior mask, query features are sent to the transformer after a merging and flattening operation. We set $m=4$ in the main experiments. All data in the following tables are the average results of five experiments with the same settings. Specific implementation details are presented in the Appendix.

\subsection{Comparison with State-of-the-Arts}
We compare the performance of RiFeNet with other classical or effective methods on $PASCAL-5^i$, as is shown in Tab.\ref{sotapascal} and Tab. A of the Appendix. RiFeNet outperforms the best method under most of the experimental scenarios. RiFeNet outperforms CyCTR by about 3.5\% under the 1-shot setting and about 2\% for the 5-shot one. Compared with the existing state-of-the-art DCAMA in the 1-shot setting with ResNet50 backbone, it surpasses by 2.5\%, rising to 2.7\% with the use of ResNet101. As for the larger gain in the 1-shot than in the 5-shot setting, we attribute this to the decreasing impact weight due to the constant amount of unlabeled data, as the ratio of unlabeled to labeled images decreases from 2 to 0.4. As labeled images increase, the positive effects of the unlabeled branch decrease, with a proportional decline in performance gain in 5-shot.

Similar experiments on COCO support the above conclusion in Tab.\ref{sotacoco}. Faced with a scenario with multiple objects in this dataset and a complex environment, RiFeNet still outperforms the current best DCAMA by 0.8\% for almost all splits in the 1-shot setting. The comparison results demonstrate the benefits of RiFeNet. The unlabeled branch provides RiFeNet with richer relevant information, which in turn improves the performance of the model. 

Qualitative results also prove the effectiveness of RiFeNet. In Fig.\ref{segresult} and Fig.A of the Appendix, the foreground objects in support and query images vary a lot, with inconsistent postures, appearances, and angles of photography. Despite this large intra-class variability, RiFeNet achieves significant improvement in maintaining foreground semantic consistency compared with the baseline. As for the similarity of background and foreground, the model deals with this binary identification much better even in cases with neighboring objects of similar appearance, with foreground occlusion, and with multiple classes of objects. Looking back to Fig.\ref{introduction}, our extracted features are essential to maintain foreground semantic consistency and provide inter-class distinction for binary classification. 

\subsection{Ablation Studies}

\textbf{RiFeNet improves the pixel-level binary classification.}
To demonstrate the effectiveness of our proposed unlabeled enhancement and multi-level prototypes in RiFeNet, we conduct diagnostic experiments in Tab.\ref{ablation}. All comparisons are set under a 1-shot setting, with ResNet50 as the backbone. Using either the unlabeled branch or multi-level prototype interaction results in a boost of approximately 2\%. When two strategies work together, RiFeNet improves by 3.1\% on top of the baseline.
\begin{figure}[!t]
\centering
\includegraphics[width=0.95\linewidth]{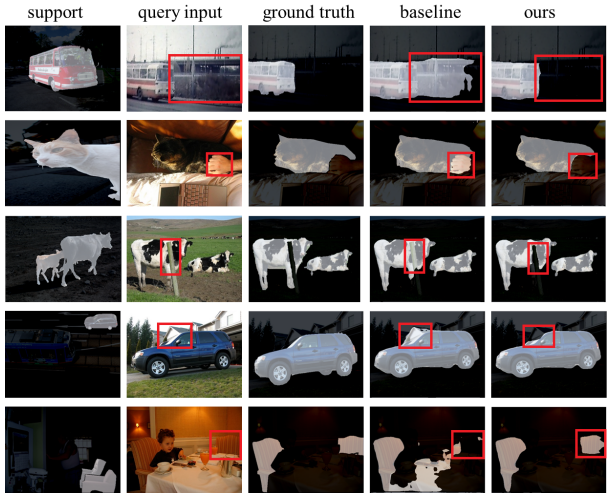}
\caption{Qualitative segmentation results on novel classes on PASCAL-$5^i$. From \textit{left} to \textit{right}: support image with mask, query input, query ground-truth mask, query prediction of the baseline, and prediction of RiFeNet.}
\label{segresult}
\end{figure}

\textbf{Different design choices of multi-level prototypes.}
We conduct ablation experiments on the model design details mentioned for the multi-level prototype, as is shown in Tab.\ref{design1}. Consistent with the theoretical analysis in Sec.3.4, it proves that our practices such as adding guidance to unlabeled branches are reasonable and reliable.

\textbf{Different design choices of unlabeled branches.}
We conduct experiments with different designed unlabeled branches to further explore their effect. As shown in Tab.\ref{design2}, the unlabeled branch without guided query prototypes results in even worse performance than the baseline, which is consistent with our analysis in Sec.3.3. On the other hand, because the unlabeled inputs come from resampling the training dataset, we double the training iterations of the baseline for a fair comparison. Increased training iterations have little effect on the baseline due to early convergence. This proves that the effectiveness of our method is not from the multiple sampling of data but from the learned discriminative and semantic features. 

\begin{table}[!t]
\centering
\begin{threeparttable}
\resizebox{0.85\linewidth}{!}{
\begin{tabular}{cc |c c c c|c}
\toprule[1pt]
Un&MP &split0 & split1 & split2 & split3 & mIoU\\
\midrule
&&65.7&71.0&59.5&59.7&64.0   \\
\checkmark&&67.3&71.8&66.2&59.2&66.1	   \\
&\checkmark&66.0&72.1&66.2&\textbf{60.4}&66.2  \\
\checkmark&\checkmark&\textbf{68.4}&\textbf{73.5}&\textbf{67.1}&59.4&\textbf{67.1}    \\
\bottomrule[1pt]     
\end{tabular}}
\end{threeparttable}
\caption{Ablation studies on the key components of RiFeNet. ``Un'' and ``MP'' denote the use of the unlabeled branch and the multi-level prototypes, respectively.}
\label{ablation}
\end{table}

\begin{table}[!]
\centering
\begin{threeparttable}
\resizebox{\linewidth}{!}{
\begin{tabular}{l |c c c c|c}
\toprule[1pt]
components&split0 & split1 & split2 & split3 & mIoU\\
\midrule
gp (support-only)&67.3&71.8&66.2&59.2&66.1\\
gp+gp&67.5&73.1&66.2&58.4&66.3\\
gp+lp (w/o CA)&68.1&73.2&66.7&59.1&66.8\\
gp+lp (w/ CA)&\textbf{68.4}&\textbf{73.5}&\textbf{67.1}&\textbf{59.4}&\textbf{67.1}    \\
\bottomrule[1pt]     
\end{tabular}}
\end{threeparttable}
\caption{Ablation studies on multi-level prototypes. ``gp'' and ``lp'' denote global and local prototypes, respectively. That is, ``gp+gp'' means extracting both query and support prototypes globally. ``CA'' refers to channel-wise attention.}
\label{design1}
\end{table}

\begin{table}[!t]
\centering
\begin{threeparttable}
\resizebox{\linewidth}{!}{
\begin{tabular}{l |c| c c c c|c}
\toprule[1pt]
components&epoch&split0 & split1 & split2 & split3 & mIoU\\
\midrule
w/o unlabel&200&66.0&72.1&66.2&\textbf{60.4}&66.2\\
w/o unlabel&400&66.5&72.4&65.5&59.5&66.0\\
un (w/o guide)&200&66.9&72.2&65.9&58.3&65.8\\
un (w/ guide)&200&\textbf{68.4}&\textbf{73.5}&\textbf{67.1}&59.4&\textbf{67.1}\\
\bottomrule[1pt]     
\end{tabular}}
\end{threeparttable}
\caption{Ablation studies on the unlabeled branch. ``w/ guide'' refers to the use of query local prototypes in the unlabeled branch for guidance, while ``w/o guide'' means using prototypes generated from the unlabeled branch itself.}
\label{design2}
\end{table}

\begin{table}[!t]
\centering
\begin{threeparttable}
\resizebox{0.75\linewidth}{!}{
\begin{tabular}{c|c c c c|c}
\toprule[1pt]
num & split0&split1&split2&split3&mIoU\\
\midrule
0	&66.0&72.1&66.2&60.4&66.2\\
1	&66.8&72.8&66.9&59.8&66.6\\
2	&\textbf{68.4}&\textbf{73.5}&\textbf{67.1}&59.4&\textbf{67.1}\\
3	&65.9&72.6&66.9&\textbf{59.8}&66.0\\
\bottomrule[1pt]     
\end{tabular}}
\end{threeparttable}
\caption{Ablation studies of different numbers of unlabeled images in the single meta-training process.}
\label{unlabelnum}
\end{table}

\textbf{Different hyper-parameters.} We first look into the effect of different numbers of unlabeled input in a single meta-training process. Tab.\ref{unlabelnum} shows the results on $PASCAL-5^i$ under a 1-shot setting, with ResNet50 as its backbone. The best results are obtained when the number of unlabeled images is set to 2. Initially, the segmentation effect of the model increased as the number of unlabeled images increased. When the number continues to increase, the accuracy decreases instead. We deem the reason is that when the effect of unlabeled enhancement counts much more than the query branch itself, the attention of feature mining may turn to the unlabeled branch, thus disturbing the query prediction. The segmentation accuracy decreases after the features are blurred. We also conduct detailed ablation experiments with other parameters, which are included in the Appendix.


\section{Conclusion}
In few-shot segmentation, traditional methods suffer from semantic ambiguity and inter-class similarity. Thus from the perspective of pixel-level binary classification, we propose RiFeNet, an effective model with an unlabeled branch constraining foreground semantic consistency. Without extra data, this unlabeled branch improves the in-class generalization of the foreground. Moreover, we propose to further enhance the discrimination of background and foreground by a multi-level prototype generation and interaction module.

\bibliography{aaai24}
\end{document}